\documentclass[lettersize,journal]{IEEEtaes}
\usepackage{amsmath,amsfonts,bm}
\usepackage{blindtext}
\usepackage{multicol}
\usepackage{subfigure}
\usepackage{pifont}
\usepackage{amsfonts}
\usepackage{caption}
\usepackage{floatrow}    
\usepackage{bm}
\usepackage{algorithmic}
\usepackage{algorithm}
\usepackage{array}
\usepackage{amsmath}
\usepackage{breqn}
\usepackage[caption=false,font=normalsize,labelfont=sf,textfont=sf]{subfig}
\usepackage{textcomp}
\usepackage{stfloats}
\usepackage{url}
\usepackage{verbatim}
\usepackage{graphicx}
\usepackage{cite}
\usepackage{comment}
\usepackage{tabularx}
\usepackage{subcaption}
\usepackage{graphicx}
\usepackage{cite}
\usepackage{amssymb,amsfonts,bm}
\usepackage{algorithmic}
\usepackage{graphicx}
\usepackage{textcomp}
\usepackage{xcolor}
\usepackage{comment}

\newcommand{\xmark}{\ding{55}}%
\newcommand{\cmark}{\ding{51}}%

\newcommand{\op}[1]{\operatorname{#1}}

\def\BibTeX{{\rm B\kern-.05em{\sc i\kern-.025em b}\kern-.08em
    T\kern-.1667em\lower.7ex\hbox{E}\kern-.125emX}}
\begin{document}

\title{FDCT: Frequency-Aware Decomposition and Cross-Modal Token-Alignment for Multi-Sensor Target Classification }

\author{SHOAIB MERAJ SAMI}
\author{MD MAHEDI HASAN}
\author{NASSER M. NASRABADI}
\member{Fellow, IEEE}
\affil{ {LCSEE} Dept., West Virginia University, WV, USA}  
\author{RAGHUVEER RAO}
\member{Fellow, IEEE}
\affil{Army Research Laboratory, MD, USA}

%% \author{FOURTH D. AUTHOR}
%% \affil{University of Colorado, Colorado, USA}
\iffalse
\receiveddate{Manuscript received XXXXX 00, 0000; revised XXXXX 00, 0000; accepted XXXXX 00, 0000..'' }
%% \accepteddate{XXXXX XX XXXX}
%% \publisheddate{XXXXX XX XXXX}

\corresp{information, e.g. {\itshape (Corresponding author: M. Smith)}. Here you may also indicate if authors contributed equally or if there are co-first authors.}

\authoraddress{Institute of Standards and Technology, Boulder, CO 80305 USA 
(e-mail: \href{mailto:author@boulder.nist.gov}{author@boulder.nist.gov}).  University, Fort Collins, CO 80523 USA (e-mail: \href{mailto:author@lamar.colostate.edu}{author@lamar.colostate.edu}).  for Metals, Tsukuba 305-0047, Japan 
(e-mail: \href{mailto:author@nrim.go.jp}{author@nrim.go.jp}).}

\editor{Mentions of supplemental materials and animal/human rights statements can be included here.}
\supplementary{Color versions of one or more of the figures in this article are available online at \href{http://ieeexplore.ieee.org}{http://ieeexplore.ieee.org}.}
\fi

\markboth{SAMI ET AL.}{FREQUENCY-AWARE DECOMPOSITION AND CROSS-MODAL ALIGNMENT FOR TARGET CLASSIFICATION}
\maketitle

%%%%%%%%% ABSTRACT
\begin{abstract}
In automatic target recognition (ATR) systems, sensors may fail to capture discriminative, fine-grained detail features due to environmental conditions, noise created by CMOS chips, occlusion, parallaxes, and sensor misalignment. Therefore, multi-sensor image fusion is an effective choice to overcome these constraints. However, multi-modal image sensors are heterogeneous, and have domain and granularity gaps. In addition, the multi-sensor images can be misaligned due to intricate background clutters, fluctuating illumination conditions, and uncontrolled sensor settings. In this paper, to overcome these issues, we decompose, align, and fuse multiple image sensor data for target classification. We extract the domain-specific and domain-invariant features from each sensor data. We propose to develop a shared unified discrete token (UDT) space between sensors to reduce the domain and granularity gaps. Additionally, we develop an alignment module to overcome the misalignment between multi-sensors and emphasize the discriminative representation of the UDT space.  In the alignment module, we introduce sparsity constraints to provide a better cross-modal representation of the UDT space and robustness against various sensor settings. We achieve superior classification performance compared to single-modality classifiers and several state-of-
the-art multi-modal fusion algorithms on four multi-sensor ATR datasets.

\end{abstract}
%%%%%%%%% BODY TEXT
\section{Introduction}
\label{sec:intro}
%Multi-sensor fusion is widely used in automatic targets, autonomous vehicles, hyper-spectral, medical images, terrain classification, sentiment analysis, and numerous others. Sometimes, one sensor can fail to capture the actual object because of technical and environmental issues. Also, the architecture and image-capturing spectrum is varied from sensor to sensor~\cite{one_sensor_fail_autonomous,one_sensor_fail_other_can_overcome_not_image}. 

Automatic Target Recognition (ATR) algorithms are essential for surveillance, homeland security, military operations, and rescue missions~\cite{transaction_multi-view_dr_nasser}. Improving the ATR algorithms could reduce casualties of human life and assets~\cite{c3ttl_dr_nasser}. The fusion of multiple sensors is employed to enhance the performance of a single-sensor classifier in ATR systems. However, these algorithms still encounter robustness and performance issues. Additionally, different sensors can capture various spectra and properties of interesting objects (i.e., military vehicles) 
~\cite{wu2017rgb,INN_Lite_transformer_cddfuse}. For instance, visible sensors can capture rich texture details, while infrared sensors excel at capturing thermal information in target images. Nevertheless, visible sensors are sensitive to changes in illumination, and thermal sensors cannot capture the texture details of military vehicles~\cite{TPAMI_VIF}. \\

In ATR applications, multi-sensor fusion is used for detection and classification purposes. Kwon et al.~\cite{Heesung_etal} utilize adaptive feature-based multi-sensor fusion for target detection. Breckon et al.~\cite{toby_etal} use codebook mapping of visual features and an adaptive background model for target detection. Laurenzis et al.~\cite{Martin_etal} employ acoustical antennas, radar, and different optical cameras to detect unmanned aerial vehicles (UAV). In this paper, we fuse visible, mid-wave, and long-wave infrared images of civilian, military vehicles, and ship targets for classification.\\

We aim to improve ATR algorithms (classifiers) by leveraging sensors' multifarious attributes and addressing the limitations inherent to individual sensors. To achieve these objectives, we perform several steps: i) efficient feature extraction, ii) reducing the domain \& granularity gaps, and iii) mitigating the misalignment between the image sensors. In the visible-infrared image fusion literature, high-frequency and low-frequency extraction is effectively used in~\cite{INN_Lite_transformer_cddfuse,cddfuse_copy}. However, in multi-sensor target classification, we are the first to implement this idea to extract the domain-invariant (low-frequency) and domain-specific (high-frequency) features.  Therefore, extracting different frequency aspects of the sensors can capture more nuanced features to improve the classification.
Moreover, images from different sensors are heterogeneous~\cite{wu2017rgb} and have domain gaps. Therefore, we unify both modalities' abstract features (concatenation of the high and low-frequency features) by a proposed shared unified discrete token (UDT) space. The UDT space not only decreases domain gaps but also reduces granularity gaps~\cite{fdt} between modalities.

Moreover, the electromagnetic interference, camera settings, and lighting conditions cause misalignment~\cite{zhao2017spindle,electromagnetic_interfernce} between the image sensors. We argue that reducing the misalignment and granularity gaps between the modalities is essential. For this purpose, we propose an alignment module to emphasize the discriminative features and reduce the presence of redundant and irrelevant features~\cite{niu2020improving}. This module improves the composite representation and enhances the cross-modal similarity~\cite{niu2020improving} between visible and infrared image pairs. The alignment module has multiple contrastive objectives, including instance-wise~\cite{mgca} and sparse cross-modal attention-wise techniques. In the sparse cross-modal attention scheme, we use sparse constraint~\cite{sparsemax_photo} to be robust against adversarial attacks~\cite{sparsemax_adversarial}. Additionally, we employ an intra-modal cluster assignment consistency regularization~\cite{mgca} to ensure control over unwanted deviations within the UDT space. In the multi-modal vision-language domain, token alignment is widely used~\cite{text_vis_ref4_icar,mgca}. To the best of our knowledge, we are the first researchers to utilize the token alignment approach for the multi-sensor ATR image classification task.
 
In this work, we design a \textbf{f}requency-aware \textbf{d}ecomposition and \textbf{c}ross-modal \textbf{t}oken-alignment (FDCT) framework for multi-sensor target classification. The FDCT algorithm extracts low-frequency features using a lite transformer~\cite{lite_transformer_lt_long_short} and high-frequency features using an invertible neural network~\cite{inn_invertible_neural_network} architectures. Then the extracted features are unified, aligned, and regularized. The resulting UDT tokens are used for target classification. We jointly optimize the feature extractor, alignment, and fusion modules of the FDCT framework in an end-to-end manner. Finally, our FDCT framework achieves better performance compared to a single modality and several state-of-the-art multi-modal fusion classifiers on four ATR datasets~\cite{dsiac, asif_mehmood1, VEDAI-dataset, VAIS_dataset}.

The major contribution of this paper is delineated as follows:
\begin{enumerate}
    \item We propose an effective FDCT framework that fuses multi-sensor images for classifying civilian and military targets. Our framework performs better than a single sensor and several multi-sensor classifiers on four test datasets.
    %Furthermore, in ATR image classification, we address the misalignment and domain gap between image sensors. 
    \item We introduce a unified discrete token space to minimize the domain and granularity gaps between the sensors. 
    \item In the FDCT framework, we propose an alignment module to reduce the misalignment between image sensors. This alignment module also captures and aligns nuance and intricate features. The comprehensive experiment shows that the alignment module improves the ATR classification performance.
    \item We propose a sparsity constraint in the alignment module to provide a better cross-modal representation of unified discrete token space and robustness against adversarial attack. We also introduce cross-modal regularization techniques in the unified discrete token space. 
\end{enumerate}

\section{Related Work}\label{sec:related_work}

\subsection{CNN and Transformer-based Image Classifiers}
In the computer vision field, CNN and transformer-based architectures are the mainstream~\cite{deit1_touvron2021training}. The CNN-based architectures existed decades ago~\cite{first_CNN_fukushima1980neocognitron,first_CNN_lecun}; however, after the introduction of AlexNet~\cite{alexnet}, CNN and its variants become more influential. To get better performance in the CNN architecture, several strategies have been introduced, e.g., i) deeper convolutional~\cite{vgg} architecture, ii) depth-wise convolution~\cite{efficientnet}, iii) channel and spatial attention~\cite{cbam,squeezeNet}, iv) deformable convolution~\cite{deformable}, v) skip connection~\cite{resnet}, vi) densely connected topology~\cite{densenet}, and vii) multiple kernel paths~\cite{googlenet}, etc. The different variants of CNN were the mainstream architectures in image classification before the invention of the vision transformer~\cite{ViT_dosovitskiy2020image}. 

In natural language processing literature, transformer-based architecture is the dominant network~\cite{swin1}. The transformer~\cite{attention_all_you_need} is an effective model for capturing the long-range dependency of sequential data. Because of the transformer's better performance in the natural language field, the researchers tried to successfully implement it in the computer vision field~\cite{ViT_dosovitskiy2020image}. The vision transformer~\cite{ViT_dosovitskiy2020image} introduced a fully transformer-based classification scheme that attained state-of-the-art performance without convolution. After the vision transformer, different strategies were implemented in the transformer-based vision architectures to achieve better performance~\cite{swin1}. The DeiT~\cite{deit1_touvron2021training} used different training strategies; the Swin Transformer~\cite{swin1} introduced the hierarchical transformer by using shifted window attention, and PVT~\cite{Pyramid_Wang_2021_ICCV} introduced the progressively shrinking pyramid strategy in the transformer-based vision architectures. Moreover, to enhance the performance and robustness of transformer-based vision architectures, several augmentation techniques~\cite{tokenmix,cutmix} were utilized in the literature. In this paper, we simultaneously utilize both CNN and transformer-based architecture for multi-sensor ATR classification.

\subsection{Fusion Algorithms}
Multi-modal fusion has achieved a noticeable performance improvement against the uni-modal sensor in diverse applications as in  classification~\cite{multimodal_servey}, semantic
segmentation~\cite{fusion_semantic}, action recognition~\cite{fusion_action_recognition}, visual question answering~\cite{fusion_visual_question}, etc. The fusion algorithm can be divided into three classes:  i) alignment, ii) aggregation, and iii) a mixture of aggregation and alignment.  Our paper employs a mixture of aggregation and alignment fusion for multi-sensor target classification. This subsection will delve into three broad areas of fusion: i) multi-modal sensor fusion, ii) visible and infrared image fusion, and iii) vision-language fusion.

\subsubsection{Multi-Modal Sensor Fusion}
Multi-modal sensor fusion finds extensive applications in enhancing the accuracy, reliability, and robustness of autonomous vehicle navigation~\cite{msf_navigation}, multi-object tracking~\cite{CVPR_multi_sensor_tracking,ICCV_multi_sensor_tracking}, hyper-spectral image classification~\cite{hang_hyperspectral_lidar}, automatic target recognition~\cite{uav_atr,msf_atr1,msf_atr_nasrabadi1,msf_nasrabadi_classification}, and terrain classification~\cite{terrain_ieee_transaction}, among other domains. Hang et al.~\cite{hang_hyperspectral_lidar} demonstrated the efficacy of two coupled CNN-based multi-modal networks for land-use and land-cover classification. The authors~\cite{remote_sensing_more_diverse_means_better} classified remote-sensing imagery sourced from diverse sensors. In multi-modal object tracking, Zhang et al.~\cite{ICCV_multi_sensor_tracking} introduced a pioneering multi-modality multi-object tracking (mmMOT) network that integrates information from image and LiDAR sensors. Their proposed network architecture encompasses a CNN, a point cloud feature extractor, and a robust fusion module. In the application of autonomous vehicles, Osep et al.~\cite{osep_cross_ref_mmmot} amalgamated RGB images, visual odometry, optional scene flow, and stereo sensor data. In the classification of naval targets~\cite{neval_atr}, infrared and electro-optical sensor data were fused using the maximum likelihood rule. Furthermore, for automatic target recognition~\cite{FLIR_and_LADAR}, forward-looking infrared and laser radar sensor data were fused using Bayesian or Dempster-Shafer combination methods.
In~\cite{video_description_attention}, the authors fused image, motion, and audio features for automatic video description. Nagrani et al.~\cite{nagrani2021attention_bottlenecks} leveraged a multi-modal bottleneck transformer to amalgamate audio and visual features. The contrastive learning-based TupleInfoNCE~\cite{fusion_contrastive} facilitated multi-modal semantic segmentation, object detection, and sentiment analysis.

\subsubsection{Visible and Infrared Image Fusion}
Visible and infrared image fusion (VIF) literature is extensive~\cite{TPAMI_VIF}, containing classical and deep learning-based methods. The pioneering algorithm in this domain was the deep Boltzmann machine~\cite{VIF_boltzman}. Subsequent, techniques such as CNNs~\cite{stdfusionnet}, auto-encoders~\cite{densefuse_autoencoder}, GANs~\cite{fusiongan}, transformers~\cite{fusion_transformer}, and denoising diffusion networks~\cite{zhao2023ddfm_diffusion, yue2023dif_diffusion} have gained dominance in the VIF literature. CNN-based VIF often employs multi-scale and multi-level feature extraction techniques.
VIF architectures based on neural architecture search, contrastive learning, and transfer learning have also become prevalent~\cite{TPAMI_VIF}. The authors of CDDFuse~\cite{INN_Lite_transformer_cddfuse} discussed the decomposition of low-frequency and high-frequency features from source images. The GAN-based VIF technique was first developed in 2019~\cite{fusiongan}, followed by numerous subsequent variations~\cite{ma2020ganmcc_GAN,fusion_infrared_gan}. Transformers are extensively utilized in VIF literature~\cite{TPAMI_VIF}, particularly in frameworks that aim to fuse both global and local information. For example, in the CGTF~\cite{cgtf_VIF} paper, authors combined the long-range dependence features of transformers with the local features of CNNs to produce high-quality fused images. Moreover, integrating VIF with other tasks, such as super-resolution~\cite{VIF_super-resolution} and image registration~\cite{VIF_registration_rfnet}, is a common trend in the literature. In this paper, we explore the idea of frequency-aware feature extraction and a decomposition loss, originating from one of the VIF literature~\cite{INN_Lite_transformer_cddfuse}.

\subsubsection{Vision-Language Fusion}
The fusion of language and vision domains is very extensive~\cite{I2T_importance_nips}. The works in~\cite{text_vis_ref1} and \cite{text_vis_ref2} employed the similarities between images and captions features.  Wei et al., in the iCAR paper~\cite{text_vis_ref4_icar}, proposed the bridging of classification and text-image alignment for zero- and few-shot classification on the ImageNet dataset. Cao et al.~\cite{text_vis_ref5_optimal_transport} utilized the optimal transport to unify the multi-modal embedding by minimizing the Wasserstein distance among the distributions of different modalities. The different types of alignment module~\cite{mgca} and shared encoder~\cite{fdt} are used for better performance in this field. In the proposed FDCT algorithm, several concepts of the multi-granularity alignment module are derived from the vision-language fusion literature~\cite{mgca}.

\section{Methodology}\label{sec:method}
\begin{figure*}[t]
  \centering
  
   \includegraphics[width=1\linewidth]{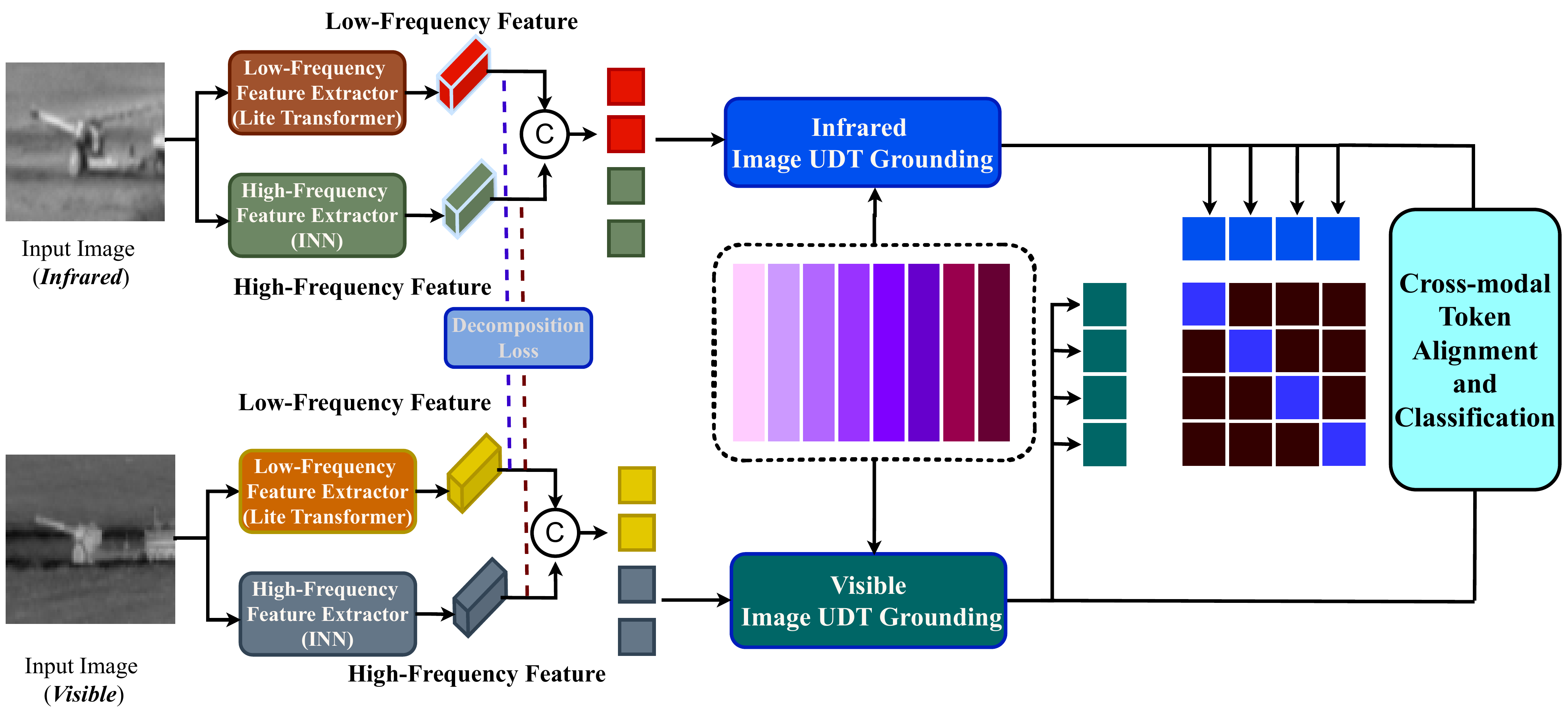}

   \caption{The proposed FDCT framework has high-frequency and low-frequency extractors across the two modalities, emphasizing the high correlation between low-frequency features and less correlation between high-frequency features. This process is pivotal for achieving modality-dependent and modality-independent feature extraction. The algorithm then passes each modality's features within a shared unified discrete token (UDT) space, reducing the granularity gap between modalities~\cite{fdt}. Furthermore, the UDT tokens are aligned and then fed to a classifier.}
   \label{fig:full_architecture1}
\end{figure*}

\begin{figure*}[t]
  \centering
  
   \includegraphics[width=1\linewidth]{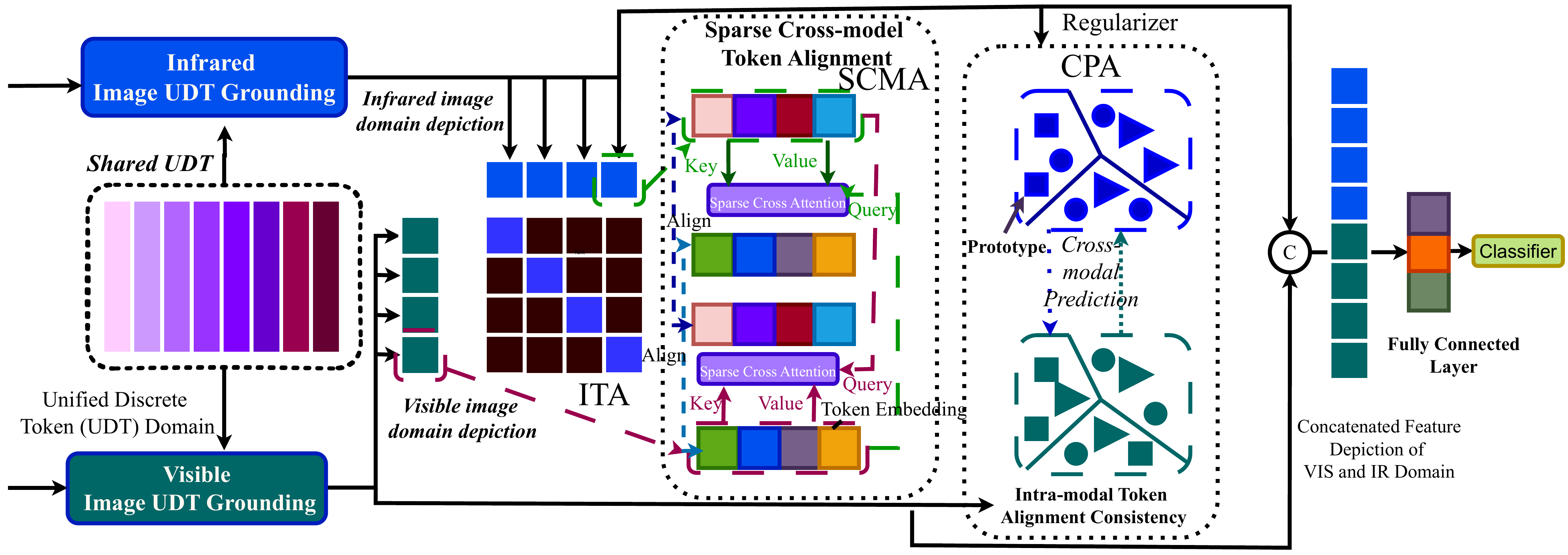}

   \caption{The illustration of the unified discrete token (UDT) space, alignment, and classification modules in the FDCT algorithm. The shared UDT tokens of the infrared and visible modalities are aligned by instance-wise cross-modal alignment (ITA), sparse cross-modal alignment (SCMA), and cross-modal prototype-level alignment (CPA) objectives. The ITA objective utilizes contrastive loss to maximize conformity among the infrared and visible modality's UDT tokens. The SCMA objective utilizes bidirectional sparse cross-attention and contrastive techniques to learn better similarity and comparison between the visible-infrared UDT tokens. The CPA objective utilizes cluster assignment and cross-modal prediction to capture the inter-instance semantic relation~\cite{mgca} between these modalities.}
   \label{fig:Full_arch_2}
\end{figure*}

\begin{figure*}[t]
  \centering
  
   \includegraphics[width=.950\linewidth]{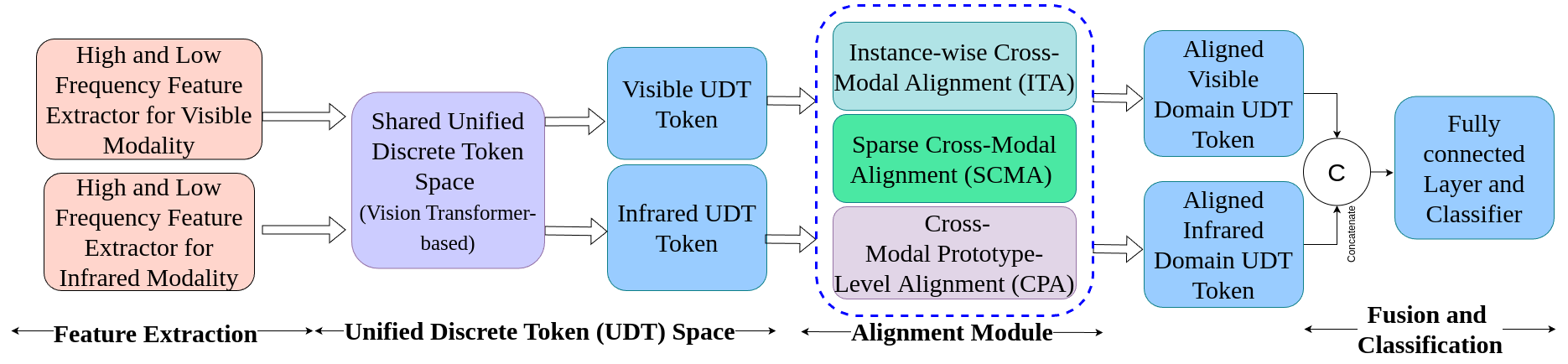}

   \caption{The block diagram of proposed FDCT framework. The low-frequency and high-frequency features of each modality are undergone through UDT space. The UDT tokens of both modalities are aligned by three cross-modal alignment objectives. The aligned tokens of both modalities are fed to a classifier. }
   \label{fig:full_block_diagram}
\end{figure*}

The proposed FDCT algorithm comprises a lite transformer network (Subsection~\ref{lt_transformer_subsection}) and an invertible neural network (Subsection~\ref{INN-sub}) for extracting low-frequency and high-frequency features. We employ a decomposition loss mechanism to ensure a high correlation between low-frequency features and a low correlation between high-frequency features across both modalities. Subsequently, each modality's low-frequency and high-frequency features are concatenated and embedded within a shared multi-modal UDT space (Subsection~\ref{UDT_sub}) to reduce the granularity gap. Due to the inherent heterogeneity of the multi-modal sensors, which leads to misalignment, we implement a trifold alignment module (Subsection~\ref{multi-granulatrity_sub}). Following alignment, the multi-modal (infrared and visible) UDT tokens are concatenated and fed to a classifier. We delineate the comprehensive architecture of the FDCT algorithm in Figures~\ref{fig:full_architecture1},~\ref{fig:full_block_diagram} and the alignment module in Figure~\ref{fig:Full_arch_2}.


\subsection{Lite Transformer}\label{lt_transformer_subsection}
\begin{figure}[t]
  \centering
  
   \includegraphics[width=0.65\linewidth]{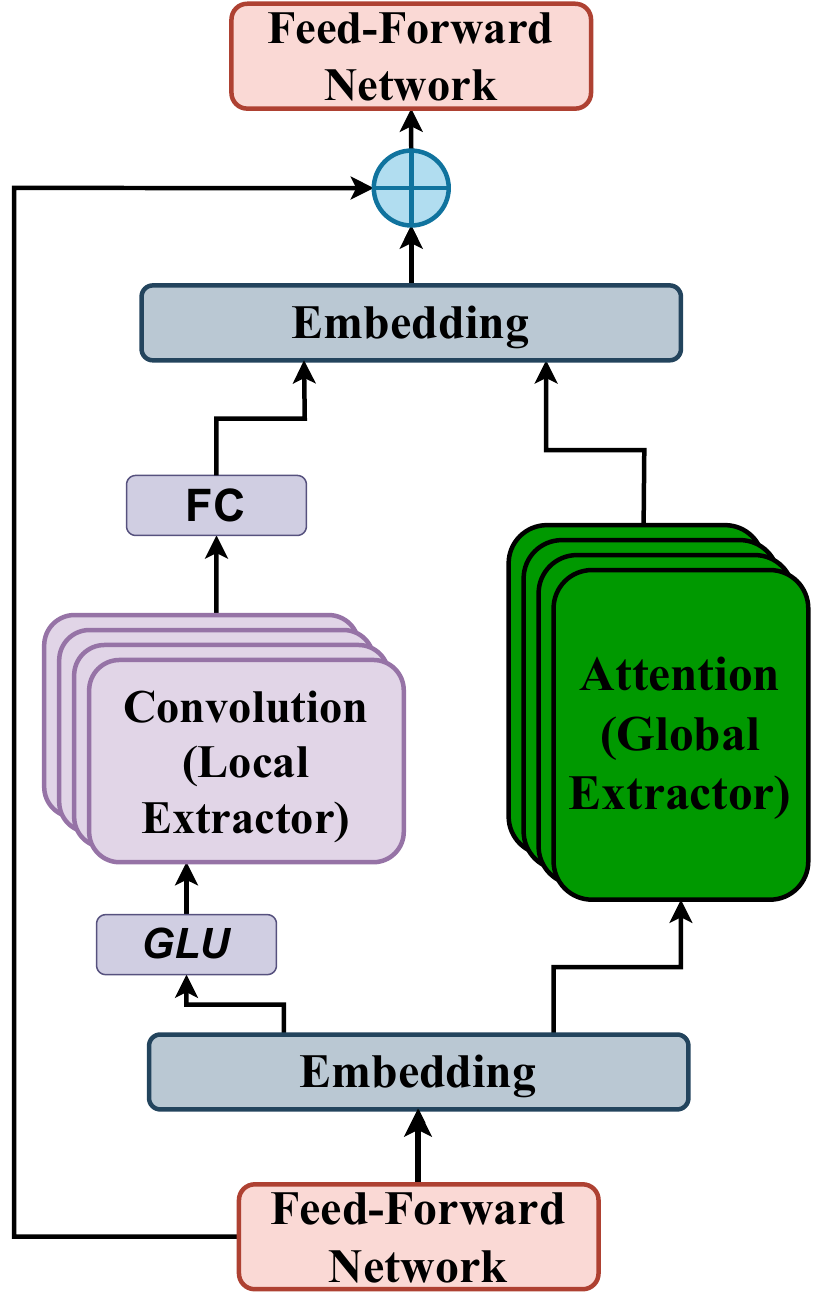}

   \caption{ The illustration of low-frequency feature extractor. The feature embeddings are split into two portions across the channel dimension. Between the two portions, one passes through a depth-wise convolution module for local context modeling, and the other portion passes through an attention module for long-distance relation extraction~\cite{lite_transformer_lt_long_short}.}
   \label{fig:lite_transformer}
\end{figure}

We utilize a lite transformer~\cite{lite_transformer_lt_long_short} as it is efficient due to computational and energy constraints. For extracting local and global features, in the lite transformer, the input features are split into two branches; one split is fed into the depth-wise convolution branch, and the other split is fed to the self-attention-based global feature extractor~\cite{lite_transformer_lt_long_short}. These processed split features are then mixed by a feed-forward block. The architecture of the lite transformer block is illustrated in~Figure~\ref{fig:lite_transformer}. We denote the lite transformer-based low-frequency feature extractor as LFE, which can extract statistical co-occurrence features from multiple image sensors. 

\begin{figure}[h]
  \centering
  
   \includegraphics[width=.99\linewidth]{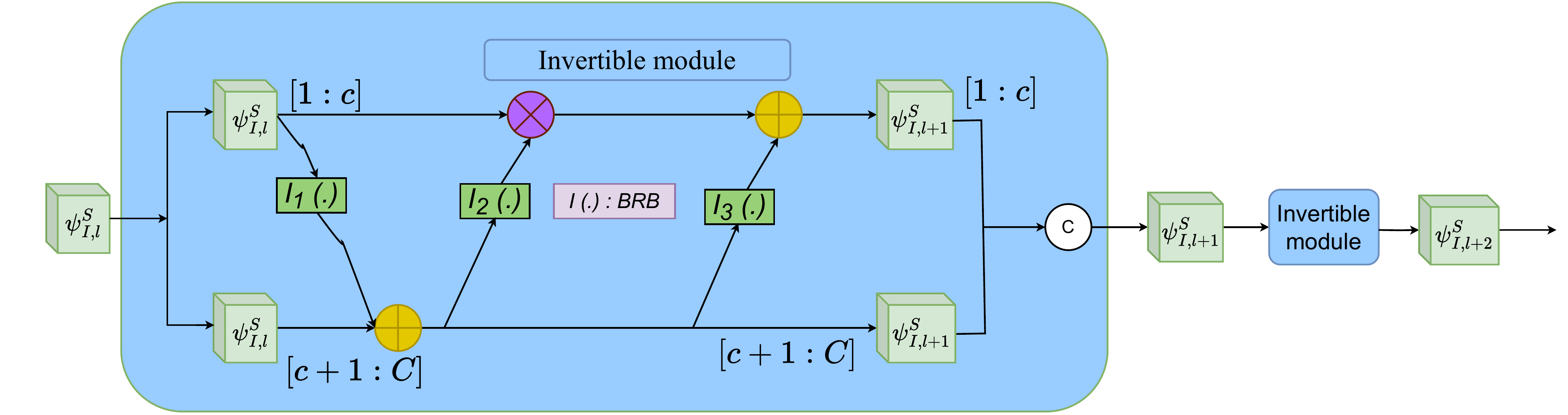}

   \caption{The architecture of high-frequency feature extractor consists of L numbers of cascaded invertible modules~\cite{INN_Lite_transformer_cddfuse}. The invertible module has an affine coupling layer consisting of scaling and translation functions and a $\odot$ Hadamard product. We use the bottleneck residual block (BRB)~\cite{mobilenetv2_BRB} to conduct the scaling and translation function. Each invertible module has three BRB blocks.}
   \label{fig:inn}
\end{figure}

\begin{figure}[h]
  \centering
  
   \includegraphics[width=1\linewidth]{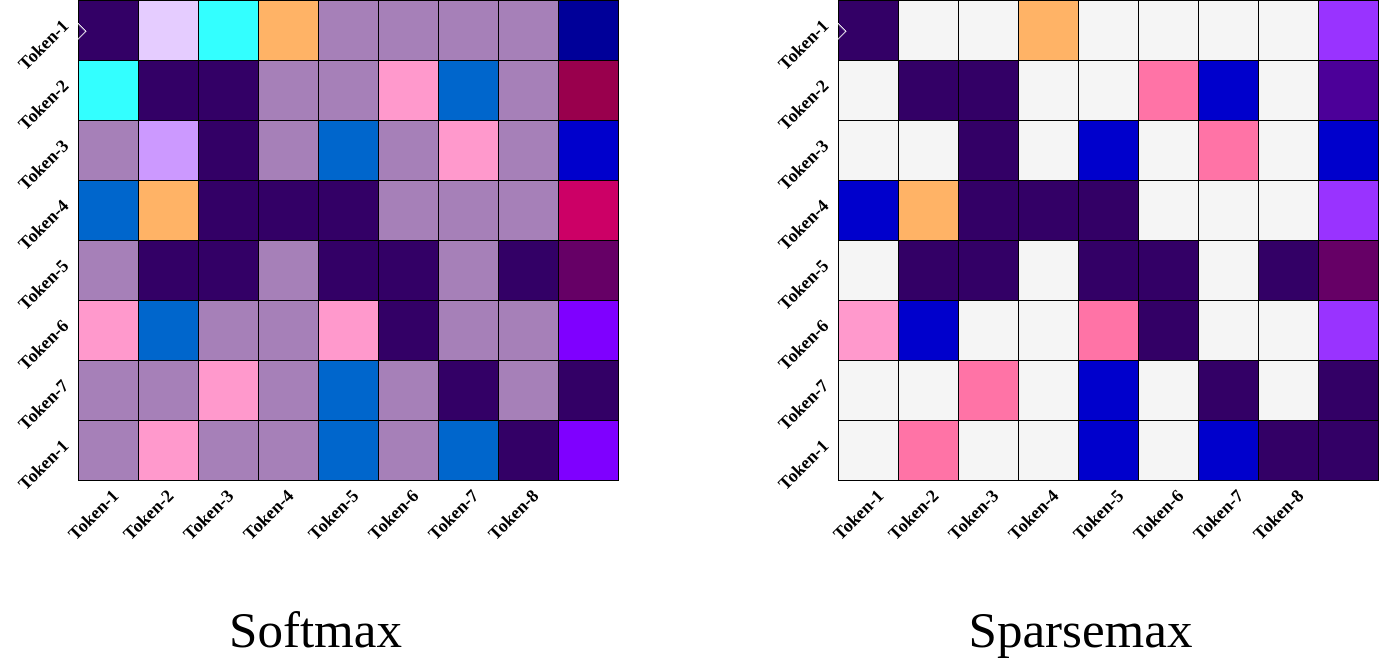}

   \caption{ The illustration of softmax and sparsemax attention wights on several tokens. The sparsemax, which first calculates a threshold, sets the weight as zero if the attention weight is less than this threshold (the white square depicts the zero attention weight)~\cite{fusedmax}.}
   \label{fig:softmax_sparsemax}
\end{figure}

\subsection{Invertible Neural Network}\label{INN-sub}

We utilize an invertible neural network (INN) to capture high-frequency detail features in our proposed FDCT algorithm. The first INN was proposed in the NICE paper~\cite{nice}. In RealNVP paper~\cite{real_nvp}, authors introduced \textit{affine coupling layer} for easier and computationally efficient data inversion. In the Glow paper~\cite{glow}, authors utilize the 1×1 invertible convolution that can yield pragmatic high-resolution images. Furthermore, INN is also applied to classification tasks for better feature extraction ability and lossless information-preserving property~\cite{invertible_classification}. 
We utilize an invertible neural network-based high-frequency feature extractor (HFE) to extract detail or high-frequency modality-independent features from an image. The extraction of high-frequency features by HFE can be expressed by:\begin{equation}\label{equ:loss_DCE}
    \psi_{I}^{HFE} = \mathcal{HF}\left(\psi_{I}^S\right), \  \psi_{V}^{HFE} = \mathcal{HF}\left(\psi_{V}^S\right),
\end{equation}
\textcolor{black}{where $\psi_{I}^{HFE}$ and $\psi_{V}^{HFE}$ represent high-frequency features extracted from the infrared and visible images, respectively. $\psi_{I}^S$ and $\psi_{V}^S$ denote the infrared and visible image pair. The function $\mathcal{HF(.)}$ denotes the high-frequency feature extraction operator.}
The HFE can extract edge and texture information by performing the mutually generated input and output features. The invertible module consists of affine coupling layers~\cite{real_nvp}. The illustration of the invertible module is in Figure~\ref{fig:inn}. In this figure, $\psi_{I,l}^{S}\left[1\!:\!c\right]$ is the first $c$ channels of the input feature at the $l$-th invertible layer, where $l=1,\cdots,L$. The arbitrary mapping functions in each invertible layer are: $\mathcal{I}_1$, $\mathcal{I}_2$, and $\mathcal{I}_3$. We utilize bottleneck residual block (BRB)~\cite{mobilenetv2_BRB} as an arbitrary mapping function in the invertible module. Moreover, $\psi_{I}^{HFE} = \psi_{I,L}^{S}$. Finally, the extraction of $\psi_{V}^{HFE}$ can be calculated in the same way as $\psi_{I}^{HFE}$.

\subsection{Unified Discrete Token Space}\label{UDT_sub}
The unified discrete token (UDT) module is a variant of vision transformer-based~\cite{ViT_dosovitskiy2020image} shared feature extractor. In the UDT module, the concatenated feature maps of the low and high-frequency feature extractors (shape $H\times W \times C$) are divided into $N$ patches. Here, $H, W, C$ are a concatenated feature map's height, width, and channel. The shape of each patch size is ($P \times P \times C$), and  $N=\frac{H \times W \times C}{P \times P \times C}$. The N patches are linearly projected into N tokens. Each token contains specific positional information; therefore, positional embedding ($\bm E_{pos}$) is also utilized. Next, the N tokens are fed to K transformer encoders, and each follows the architecture described in~\cite{attention_all_you_need}. The transformer block includes a multi-head self-attention, MLP, and layer normalization. 

\textcolor{black}{The self-attention can be denoted as:}
\begin{equation}\label{equ:self_attention}
   \mathrm{Attn}({Q}, {K}, {V}) = \mathrm{softmax}(\frac{{Q}{K}^T}{\sqrt{d_k}}){V}, 
\end{equation}
\textcolor{black}{where ${Q}$, ${K}$, ${V}$ are query, key, and value, respectively.}
The multi-head self-attention $\mathrm{MH{\text -}SA}({Q}, {K}, {V})$ can be denoted as:  
\begin{equation}\label{equ:msa}
    \mathrm{MH{\text -}SA}(\mathbf{Q}, \mathbf{K}, \mathbf{V}) 
    = \mathcal{CAT}(\mathrm{H{-}Attn_1}, \dots, \mathrm{H{-}Attn_h}) W^O,
\end{equation}
\noindent where $\mathrm{H{-}Attn}_i = \mathrm{Attn}(\mathbf{QW}^Q_i, \mathbf{KW}^K_i, \mathbf{VW}^V_i)$,
 $h$ is the number of heads in~Eq.~\ref{equ:msa}, and $\mathcal{CAT}$ represents the concatenation operation,
\textcolor{black}{and $W^O$, $W_i^Q$, $W_i^K$, $W_i^V$ are linear projection matrices. The $x_p^1$ patch is projected to $D$ dimensional vector. This flattening and projection operation is performed on all the N patches using the following equation:}
\begin{equation}\label{eq:UDT1}
\begin{split}
    \bm \iota_0 = [  \, \bm x^1_p \bm E;  \bm x^2_p \bm E; \cdots; \, \bm x^{N}_p \bm E ],\qquad 
    \bm E \in \mathbb{R}^{(P^2 \cdot C) \times D},
\end{split}
\end{equation}
where $\bm \iota_0$ is the N initial tokens.
\\
\noindent The transformer block is explained by~Eq.~\ref{eq:UDT2}:
\begin{equation}\label{eq:UDT2}
\begin{split}
    \bm \iota^\prime_\jmath = \op{MH{\text -}SA}(\op{LN}(\bm \iota_{\jmath-1})) + \bm \iota_{\jmath-1},  \qquad \jmath=1\ldots K \quad \\
    \bm \iota_\jmath = \op{MLP}(\op{LN}(\bm \iota^\prime_{\jmath})) + \bm \iota^\prime_{\jmath}, \qquad \qquad \quad  \jmath=1\ldots K  \quad \\
     \bm \tilde{\bm \iota} = \op{LN}(\bm \iota_K),~   \qquad \qquad \qquad \qquad \qquad \qquad \quad \quad \quad 
\end{split}
\end{equation}
\textcolor{black}{where $\bm \tilde{\bm \iota}$ denotes the finalized UDT tokens.\\
Let $\mathcal{D}$ comprising $M$ visible-infrared pairs, where $\mathcal{D}=\{(S_{V,1}, S_{I,1}), (S_{V,2}, S_{I,2}), \ldots, (S_{V,M}, S_{I,M})\}$. The $j^{th}$ visible-infrared image pair can be denoted as $(S_{V,j}, S_{I,j})$, which undergoes encoding through LFE, HFE, and UDT encoders. These processes generate two sets of outputs for each pair: i) a sequence of encoded visible modality tokens $\mathbf{Y}_j = \{\mathbf{y}_j^1, \mathbf{y}_j^2, \ldots, \mathbf{y}_j^N\}$ of the $j^{th}$ visible sample $\mathbf{S}_{V,j}$, ii) a sequence of infrared modality tokens $\mathbf{Z}_j = \{\mathbf{z}_j^1, \mathbf{z}_j^2, \ldots, \mathbf{z}_j^N\}$  of the $j^{th}$ infrared sample $\mathbf{S}_{I,j}$. Here, $N$ is the number of UDT tokens for each sample. Subsequently, the UDT tokens of the infrared and visible pairs undergo processing through the alignment module.}
\subsection{Multi-Granularity Cross-Modal Alignment}\label{multi-granulatrity_sub}
The purpose of the alignment module is to align the visible and infrared image representations for target classification. The multi-granularity cross-modal alignment module consists of three ways of aligning, i.e., i) instance-wise, ii) cross-modal token-wise, and iii) prototype-level~\cite{mgca}.
Figure~\ref{fig:Full_arch_2} illustrates the multi-granularity cross-modal alignment module between visible and infrared sensor images.
We utilize an instance-wise token alignment (ITA)~\cite{mgca}  scheme to maximize the concurrence between visible-infrared pairs. Moreover, we develop a sparse cross-modal alignment (SCMA) scheme to maximize the fine-grained resemblance between visible-infrared tokens. Finally, the inter-instance association is preserved by the cross-modal prototype alignment (CPA)~\cite{mgca} scheme.
\paragraph{Instance-wise Cross-Modal Alignment}
In our proposed FDCT algorithm, we employ an instance-wise token alignment (ITA)~\cite{mgca} scheme to facilitate the algorithm to correctly map the visible-infrared image pairs close to the latent space while projecting arbitrary pairs farther away. 
For the \(j\)-th visible-infrared UDT tokens \(\mathbf{y}_j\) and \(\mathbf{z}_j\) undergo through two non-linear projection function (\(h_V\) and \(h_I\)). These projection functions transform \(\mathbf{y}_j\) and \(\mathbf{z}_j\) into low-dimensional embeddings \(\hat{\mathbf{y}}_j\) and \(\hat{\mathbf{z}}_j\), respectively.
Furthermore, the cosine similarity of $j^{th}$ visible-infrared token pair can be expressed by the following equation:
\begin{equation}
    \label{eq:1}
    sim(S_{V,j}, S_{I,j}) = \hat{\mathbf{y}}_j^T\hat{\mathbf{z}}_j,
\end{equation}
$\mathrm{where}\ \hat{\mathbf{y}}_j=h_V(\mathbf{y}_j), \mathrm{and}\ \hat{\mathbf{z}}_j=h_I(\mathbf{z}_j)$.
We utilize two symmetric InfoNCE losses (visible-to-infrared token pair and infrared-to-visible token pair) to hold the joint information among the true pairs in the lower-dimensional embedding space. The following equations can express these InfoNCE losses:
\begin{dmath}
\ell_j^{\mathrm{V2I}} = -\mathrm{log}\frac{\mathrm{exp}(sim(S_{V,j}, S_{I,j}) / \gamma_1)}
    {\sum_{k=1}^B\mathrm{exp}(sim(S_{V,j}, S_{I,k})/\gamma_1)}
    ,  
\end{dmath}
\begin{dmath}
\ell_j^{\mathrm{I2V}} = -\mathrm{log}\frac{\mathrm{exp}(sim(S_{I,j}, S_{V,j}) / \gamma_1)}{\sum_{k=1}^B\mathrm{exp}(sim(S_{I,j}, S_{V,k})/\gamma_1)},
\end{dmath}
\noindent where $B$ is the batch of tokens and $\gamma_1$ is the temperature hyperparameter of the ITA.
The instance-wise token alignment (ITA) loss function can be expressed by:
\begin{equation}
\mathcal{L}_{\mathrm{ITA}} = \frac{1}{2M}\sum_{j=1}^{M}(\ell_j^{V2I} +\ell_j^{I2V}),
\end{equation}
where $M$ is the total number of visible-infrared image pairs.
% where $B$ is the batch size. 
\paragraph{Sparse Cross-Modal Alignment}
Detail information is crucial in multi-sensor automatic target classification as it often contains class-specific discriminative properties, even within a small patch of the target chip. However, the ITA objective may overlook significant intricate clues between visible-infrared image pairs. To capture and align the nuance of mutual information between the sensors, we propose a sparse attention-based bidirectional token-wise cross-modal alignment (SCMA) scheme incorporating sparsemax-based attention~\cite{sparsemax_photo}. This attention function calculates a threshold, and below this threshold weight becomes zero. Both softmax and sparsemax attention are depicted in Figure~\ref{fig:softmax_sparsemax}. The SCMA scheme bi-directionally compares, aligns, and correlates between cross-modal representations of infrared and visible image pairs.
In particular, for the $j^{th}$ visible-infrared image pair $(S_{V,j}, S_{I,j})$, the projected visible and infrared  UDT tokens ($\mathbf{Y}_j$ and $\mathbf{Z}_j$) are  projected onto a normalized lower-dimensional embedding, these can be expressed as: 
$\tilde{\mathbf{Y}}_j = \{\tilde{\mathbf{y}}_j^1, \tilde{\mathbf{y}}_j^2, ..., \tilde{\mathbf{y}}_j^N\}$ 
and
$\tilde{\mathbf{Z}}_j = \{\tilde{\mathbf{z}}_j^1, \tilde{\mathbf{z}}_j^2, ..., \tilde{\mathbf{z}}_j^N\}$, 
here, the dimension of $\tilde{\mathbf{y}}_j \in \mathbb{R}^{d}$ and $\tilde{\mathbf{z}}_j \in \mathbb{R}^{d}$.
In the SCMA scheme, we initially match the visible and infrared image tokens with a cross-attention scheme. For this purpose, the $i^{th}$ UDT token of $j^{th}$ visible image $\tilde{\mathbf{y}}_j^i$ attend to all infrared token embedding in $\tilde{\mathbf{Z}}_j$. In this respect, we calculate corresponding cross-modal sparse attention embedding $\mathbf{a}_j^i$:
\begin{equation}\label{sparse_wt_eq}
\resizebox{1\hsize}{!}{
    $\mathbf{a}_j^i = \sum_{k=1}^N O[\alpha_j^{i2k}[\hat{V}\tilde{\mathbf{z}}_j^k]],\,
    \alpha_j^{i2k} = \mathrm{sparsemax}[\frac{(\hat{Q}\tilde{\mathbf{y}}_j^i)^{T}(\hat{K}\tilde{\mathbf{z}}_j^k)}{\sqrt{d}}],$}
\end{equation}
where the dimension $\hat{Q}$, $\hat{K}$, and  $\hat{V}$ is $\mathbb{R}^{d \times d}$. Then, we utilize a visible-infrared token alignment (VITA) loss $\mathcal{L}_{\mathrm{VITA}}$ to pull the visible token $\tilde{\mathbf{y}}_j^i$ close to its cross-modal attention embedding $\mathbf{a}_j^i$, but $\tilde{\mathbf{y}}_j^i$ is pushed away from other cross-modal attention embeddings.
Moreover, different visible tokens have different importance; therefore, we utilize a weight $w_j^i$ while computing the VITA loss. The VITA loss can be denoted as:
\begin{equation}
\begin{split}
\resizebox{.99\hsize}{!}{
    $\mathcal{L}_{\mathrm{VITA}} = -\frac{1}{2MN}\sum_{j=1}^M \sum_{i=1}^{N} w_j^i
    (\mathrm{log}\frac
    {\mathrm{exp}(sim(\tilde{\mathbf{y}}_j^i, \mathbf{a}_j^i)/\gamma_2)}
    {\sum_{k=1}^S \mathrm{exp}(sim(\tilde{\mathbf{y}}_j^i, \mathbf{a}_j^k)/\gamma_2)}$}  \\
   \resizebox{.5\hsize}{!}{ $+ \mathrm{log}\frac
    {\mathrm{exp}(sim(\mathbf{a}_j^i, \tilde{\mathbf{y}}_j^i)/\gamma_2)}
    {\sum_{k=1}^S \mathrm{exp}(sim(\mathbf{a}_j^i, \tilde{\mathbf{y}}_j^k)/\gamma_2)}
    ),$}
\end{split}
\end{equation}
\noindent where $\gamma_2$ is the temperature hyperparameter of the SCMA.
Likewise, for the $i^{th}$ infrared UDT token $\tilde{\mathbf{z}}_j^i$ in $j^{th}$ infrared image, we also compute the attention embedding $\hat{\mathbf{a}}_j^i$. The infrared-visible token alignment (IVTA) loss function is calculated similar to VITA. The total objective function of the proposed SCMA objective is the summation of VITA and IVTA losses: 
\begin{equation}
    \mathcal{L}_{\mathrm{SCMA}} = \frac{1}{2}(\mathcal{L}_{\mathrm{VITA}} + \mathcal{L}_{\mathrm{IVTA}}).
\end{equation}
Notably, the difference between SCMA and ITA is that SCMA considers cross-attention and sparsity constraints when calculating InfoNCE loss.
\paragraph{Cross-Modal Prototype-Level Alignment} The ITA and SCMA objectives consider negative token pairs if two tokens come from different positions. This phenomenon is sometimes too strict because many tokens share common high-level semantic embedding. Moreover, this phenomenon sometimes wrongly contrasts between negative pairs. To overcome this issue, we utilize a cross-modal prototype alignment (CPA)~\cite{mgca} objective to regularize and properly align the cross-modal inter-instance agreement between the unified discrete token space of image pairs. We utilize the iterative Sinkhorn-Knopp clustering algorithm~\cite{cuturi2013sinkhorn} in the UDT token pair $(\hat{\mathbf{y_j}}, \hat{\mathbf{z_j}})$ (derived from Eq. \ref{eq:1}). This clustering algorithm assigns two soft cluster task identifiers $\mathbf{q}_{V,j}$  and $\mathbf{q}_{I,j}$, by separately assigning $\hat{\mathbf{y_j}}$ and $\hat{\mathbf{z_j}}$ into $E$ distinct clusters, $\mathcal{C} = \{\mathbf{c}_1, ..., \mathbf{c}_E\}$. Here, $\mathcal{C}$ is $E$ trainable cross-modal cluster instance set. The dimension of $\mathbf{c}_e$ is $\mathbb{R}^d$.
After that, we compute the visible image UDT token softmax probability ($\mathbf{p}_{V,j}$) between $\hat{\mathbf{y_j}}$ and all the prototype clusters in $\mathcal{C}$. 
\begin{equation}
    \mathbf{p}_{V,j}^{(e)} = \frac{\mathrm{exp}(({{\mathbf{\hat{y_j}})}}^T\mathbf{c_e}/\gamma_3)}
    {\sum_{e'}\mathrm{exp}(({\mathbf{\hat{y_j}}})^T\mathbf{c_{e'}}/\gamma_3)}.
\end{equation}
In a similar manner, we also calculate infrared image UDT token softmax probability as $\mathbf{p}_{I,j}$: 
\begin{equation}
    \mathbf{p}_{I,j}^{(e)} = \frac{\mathrm{exp}(({\mathbf{\hat{z_j}})^T\mathbf{c_e}/\gamma_3)}}
    {\sum_{e'}\mathrm{exp}((\mathbf{\hat{z_j}})^T\mathbf{c_{e'}}/\gamma_3)},
\end{equation}
\noindent where $\mathbf{p}_{I,j} \in \mathbb{R}^E$. Also $\mathbf{p}_{I,j}$ is the cosine similarities between ${\mathbf{\hat{z_j}}}$ and all the cross-modal prototype clusters in ($\mathcal{C}$). $\gamma_3$ is the temperature hyperparameter of the CPA, and `(e)' denotes the $e$-th element of the cluster assignment vector.
The cross-modal prototype alignment is accomplished by employing cross-entropy loss between the \textit{softmax prediction ($\mathbf{p}_{V,j}^{(e)}$) } and \textit{soft cluster task identifier} ($\mathbf{q}_{I,j}^{(e)}$) as shown below:
\begin{equation}
\ell(\mathbf{\hat{y_j}}, \mathbf{q}_{I,j}) = \sum_{e=1}^E\mathbf{q}_{I,j}^{(e)}\mathrm{log} \, \mathbf{p}_{V,j}^{(e)},
\end{equation}
\begin{equation}
    \ell(\mathbf{\hat{z_j}}, \mathbf{q}_{V,j}) = \sum_{e=1}^E\mathbf{q}_{V,j}^{(e)}\mathrm{log}\, \mathbf{p}_{I,j}^{(e)}.
\end{equation}
The total CPA objective can be expressed as:
\begin{equation}
    \mathcal{L}_{\mathrm{CPA}} = \frac{1}{2M}\sum_{j=1}^M (\ell(\mathbf{\hat{y_j}}, \mathbf{q}_{I,j}) + \ell(\mathbf{\hat{z_j}}, \mathbf{q}_{V,j})).
\end{equation}
\subsection{Decomposition Loss and Categorical Cross-Entropy Loss}\label{decompositon_CE_loss}
In this Subsection, we will describe the decomposition loss of the low-frequency and high-frequency feature extractors. Furthermore, we also describe the cross-entropy loss of the classifier. 

\paragraph{Decomposition Loss of LFE and HFE} The decomposition loss ($\mathcal{L}_{decp}$) can be denoted as:
\begin{equation}\label{equ:loss_decomposition}
\resizebox{.88\hsize}{!}{
$\mathcal{L}_{decp} = \frac{\left(\mathcal{L}_{Correlation}^{HFE}\right)^2}{{\mathcal{L}_{Correlation}^{LFE}+\beta}} =\frac{\left({Cor{-}coff}\left(\psi_{V}^{HFE},\psi_{I}^{HFE}\right)\right)^2}{{Cor{-}coff}\left(\psi_{V}^{LFE},\psi_{I}^{LFE}\right)+\beta},$}
\end{equation}
where ${Cor{-}coff({., .})}$  is the correlation coefficient operator, and $\beta$ is set to 1.0001 to confirm that the decomposition loss term always remains as a positive value.\\
The low-frequency features extracted from the visible and infrared images contain background elements and broad-scale contextual information that strongly correlate with the domain-shared attributes. The decomposed low-frequency features of the infrared and visible pair, denoted as $\psi_{I}^{LFE}$ and $\psi_{V}^{LFE}$, have a strong correlation. Thus, we expect the correlation loss between the low-frequency features ($\mathcal{L}_{Correlation}^{LFE}$) to have a larger value. On the other hand, the high-frequency features ($\psi_{V}^{HFE}$ and $\psi_{I}^{HFE}$), comprising texture details from visible and infrared images, depict domain-specific attributes that exhibit less correlation between pairs. Hence, the correlation between the high-frequency features ($\mathcal{L}_{Correlation}^{HFE}$) is expected to approach insignificance. Therefore, the numerator and denominator of decomposition loss ($ \mathcal{L}_{decp}$) contains $\mathcal{L}_{Correlation}^{HFE}$ and $\mathcal{L}_{Correlation}^{LFE}$, respectively.

% \subsection{Technical detail}
\paragraph{Cross-Entropy loss of the classifier}
 In the FDCT framework, we optimize the categorical cross-entropy loss ($\mathcal{L}_{Cross-En}$) of the classifier. The classifier's cross-entropy loss encourages the network to learn, discriminate, and generalize multi-sensor image representation. The $\mathcal{L}_{Cross-En}$ can be denoted as:

\begin{equation}\label{eq:cross-entropy}
  \begin{aligned}
   \mathcal{L}_{Cross-En} = -\sum_{cls=1}^{CLS}y_{out,cls}\log(p_{out,cls}),
    \end{aligned}
\end{equation}
\noindent where $y_{out,cls}$ and $p_{out,cls}$ denote the true label and the predicted probability, respectively, of the ATR training data. The class number is denoted by `cls'.
\subsection{Total Objective}\label{overall_obj}
During the training of FDCT, we jointly optimize the three alignment objectives, the feature extractor's decomposition objective, and the classifier's cross-entropy objective. The overall objective can be described as follows: 
\begin{equation}\label{eq:total_obj}
    \begin{aligned}
        \mathcal{L}_{Total-Loss} &= \sigma_1 * \mathcal{L}_{\mathrm{ITA}} + \sigma_2 * \mathcal{L}_{\mathrm{SCMA}} + \sigma_3 * \mathcal{L}_{\mathrm{CPA}} \\
        &+\Gamma_1 *\mathcal{L}_{decp} +\Gamma_2 *\mathcal{L}_{Cross-En} ,
    \end{aligned}
\end{equation}
where $\sigma_1$, $\sigma_2$, and $\sigma_3$ are the hyperparameters to balance the trifold alignment forces. Furthermore, $\Gamma_1$ and $\Gamma_2$ are the decomposition and cross-entropy loss hyperparameters, respectively.

\section{Experiments}\label{sec:experiments}

\subsection{Dataset}
We experiment the proposed FDCT framework on four ATR datasets. The DSIAC~\cite{dsiac}, FLIR ATR~\cite{asif_mehmood1}, VAIS~\cite{VAIS_dataset}, and VEDAI~\cite{VEDAI-dataset} datasets consist of paired ATR vehicle images that are captured at the same time by multiple sensors in different environmental scenarios.

\begin{figure*}[!b] 
\centering
 \makebox[\textwidth]{\includegraphics[height=1.07\paperwidth]{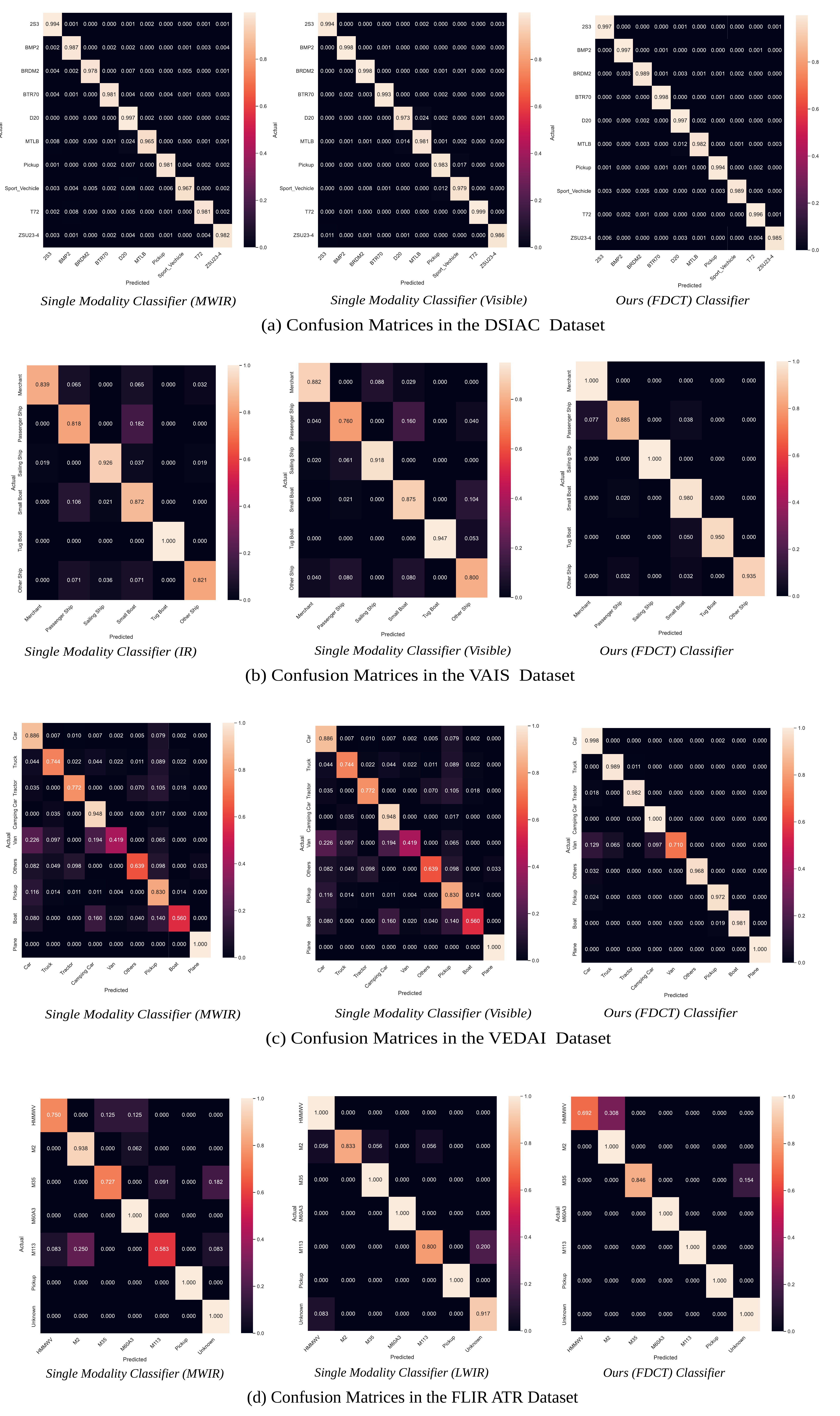}}
\caption{The confusion matrices for the single modality and multi-modal fusion (FDCT) classifiers using the (a) DSIAC, (b) VAIS, (c) VEDAI, and (d) FLIR ATR datasets.}
\label{fig:Fig_combined_confusion}
\end{figure*}

\begin{figure*}[!b] 
\centering
 \makebox[\textwidth]{\includegraphics[height=1.02\paperwidth]{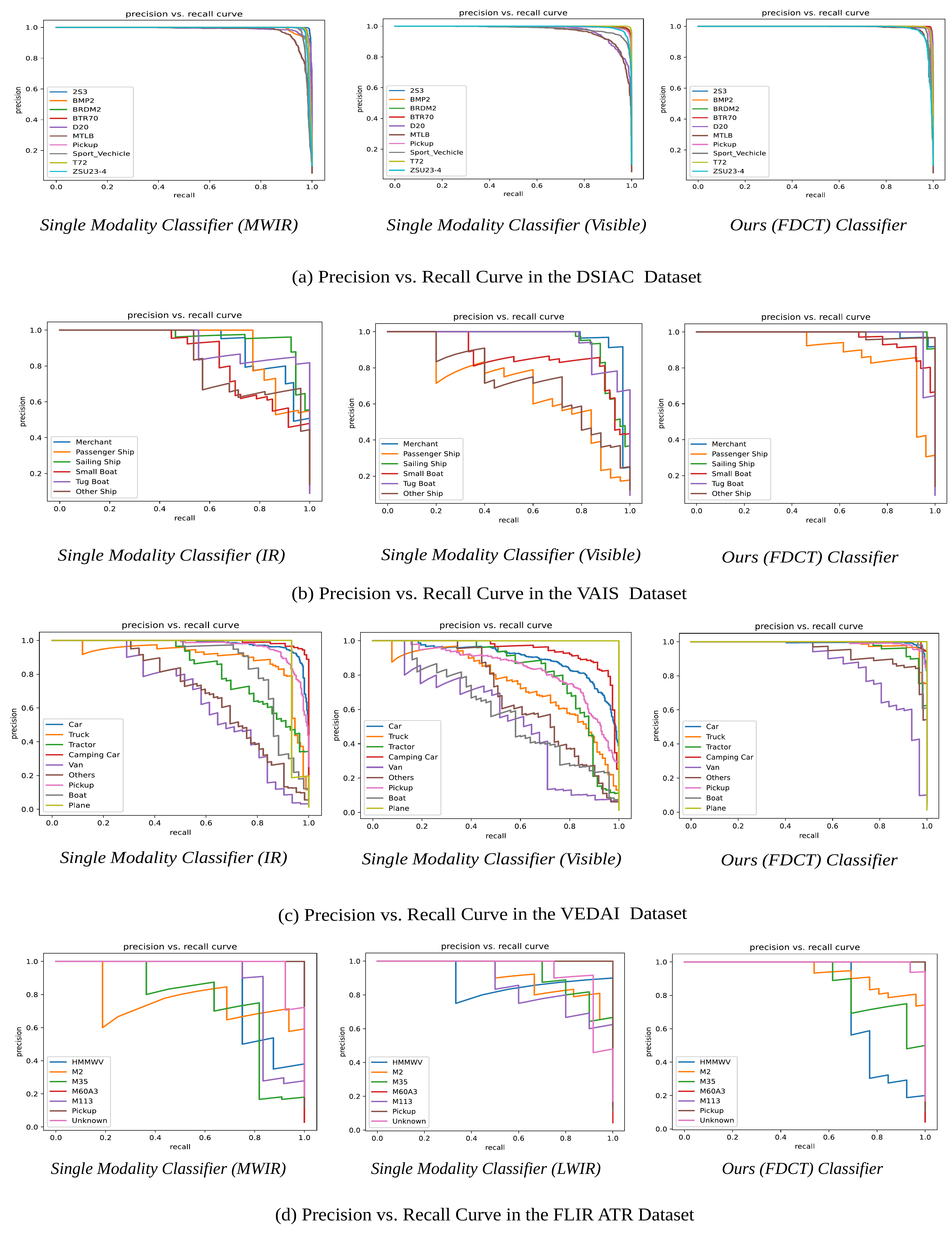}}
\caption{The precision vs. recall curves for the single modality and multi-modal fusion (FDCT) classifiers using the (a) DSIAC, (b) VAIS, (c) VEDAI, and (d) FLIR ATR datasets.}
\label{fig:Fig_combined_PR}
\end{figure*}

\subsubsection{DSIAC dataset}
%Our fusion framework, FDCT, is implemented within 

The publicly available DSIAC dataset~\cite{dsiac} has ten distinct vehicle classes captured in both visible and mid-wave infrared (MWIR) modalities. The ten classes of vehicles are: `Pickup', `Sport vehicle', `D20', `2S3', `BTR70', `BRDM2', `BMP2', `MT-LB', `T72', and `ZSU23-4'~\cite{c3ttl_dr_nasser}.  The visible domain has 340,200, and the MWIR domain has 174,600 image frames. In our research, we detect target chips using a detector~\cite{meta-uda}. 

\subsubsection{FLIR ATR Dataset}
In assessing the efficacy of our proposed FDCT framework, we leverage mid-wave (MW) and long-wave (LW) images of FLIR ATR dataset~\cite{asif_mehmood1}. This dataset comprises 461 paired MW and LW infrared images~\cite{c3ttl_dr_nasser}. This dataset consists of seven classes of target vehicles: `PICKUP, `M2', `HMMWV', `M60A3', `M35', `M113', and `UNKNOWN'. 

\subsubsection{Visible and Infrared Spectrums (VAIS) Dataset}
The VAIS dataset~\cite{VAIS_dataset} consists of 1,242 infrared and 1,623 visible modality maritime targets of 1,088 corresponding pairs. This dataset encapsulates six major categories: `Tugboat', `Small Boat',  `Medium Other Ship' `Merchant Ship',  `Sailing Ship',  and `Medium Passenger Ship'.

\subsubsection{VEDAI Dataset}

The VEDAI dataset~\cite{VEDAI-dataset} consists of vehicle targets for aerial imagery. The VEDAI dataset contains nine target classes, which are: `Boat', `Camping Car', `Car', `Others', `Pickup', `Plane', `Tractor', `Truck', and `Vans'. The VEDAI dataset incorporates multi-sensor images of infrared and visible modalities.

The total distribution of the VAIS, VEDAI, DISAC, and FLIR ATR datasets is available in the~\cite{c3ttl_dr_nasser} article. We divide the training, testing, and validation sets by randomly partitioning each dataset~\cite{dsiac,asif_mehmood1,VAIS_dataset,VEDAI-dataset} into 70:20:10 ratios, respectively.

\subsection{Training Details}

To train the proposed FDCT framework, we use the AdamW optimizer~\cite{adamw} with a weight decay of 0.05, a learning rate of $2e{-4}$. We use a cosine annealing scheduler~\cite{cosine_annealing}. For the alignment module, the temperature hyperparameters are $\gamma_1=0.1$, $\gamma_2=0.07$, and $\gamma_3=0.2$. We assigned the number of prototypes to $K=10$. The total number of UDT tokens for each image is $N=196$. We set $\sigma_1=1$, $\sigma_2=1$, $\sigma_3=1$, $\Gamma_1=1$, and $\Gamma_2=1$ for the overall objective (Eq.~\ref{eq:total_obj}). We train the FDCT framework for 30 epochs on the DSIAC dataset and 70 epochs on the remaining three datasets. \\
We set ResNet-50~\cite{resnet} as a single modality classifier. To train the single modality classifier, we utilize AdamW~\cite{adamw} optimizer with the learning rate of $2 e{-4}$ and a cosine annealing scheduler. Moreover, we train the single modality classifier for 20, 100, 100, and 70 epochs on the DSIAC, VAIS, VEDAI, and FLIR ATR datasets, respectively.  We conduct these experiments on two NVIDIA RTX-6000 Ada GPUs by utilizing the PyTorch framework~\cite{pytorch}. All the images are resized to $224\times 224 \times 3$, and we set the batch size as 24.

\begin{table}
\small
%\tablestyle
\caption{ Performance of different classifiers on the DSIAC dataset.} 
\label{tab:different_class_dsiac}
\begin{tabular}{ll} %% this 
\hline
Classifier Name  & Accuracy (\%)\\ \hline
Single modality \footnotesize(MWIR) &98.09 \\ 

Single modality {\footnotesize(Visible)} &98.67\\ 

\textbf{FDCT }&\textbf{99.29}\\
\hline
{\footnotesize Comparison with other SOTA methods} &  \\ \hline
TFormer~\cite{sota_tformer} &99.25\\
GeSeNet~\cite{sota_gesenet}   &99.09\\
FusionM4Net~\cite{sota_fusionm4net} & 99.03\\

\hline
\end{tabular}
\end{table} 

\begin{table}
\small
%\tablestyle
\caption{Performance of different classifiers on the VAIS dataset.} 
\label{tab:different_class_VAIS}
\begin{center}       
\begin{tabular}{ll} %% this 
\hline
Classifier Name  & Accuracy (\%)\\ \hline
Single modality \footnotesize(IR) &88.00\\ 

Single modality {\footnotesize(Visible)} &86.98\\ 

\textbf{FDCT }&\textbf{96.82}\\
\hline
{\footnotesize Comparison with other SOTA methods} &  \\ \hline

TFormer~\cite{sota_tformer} &96.63\\
GeSeNet~\cite{sota_gesenet}   &88.62\\
FusionM4Net~\cite{sota_fusionm4net} & 96.15\\

\hline

\end{tabular}
\end{center}
\end{table} 

\begin{table}
\small
%\tablestyle
\caption{Performance of different classifiers on the VEDAI dataset.} 
\label{tab:different_class_VEDAI}
\begin{center}       
\begin{tabular}{ll} %% this 
\hline
Classifier Name  & Accuracy (\%)\\ \hline
Single modality \footnotesize(IR) &86.11\\ 

Single modality {\footnotesize(Visible)} &82.29\\ 

\textbf{FDCT}&\textbf{97.97}\\
\hline
{\footnotesize Comparison with other SOTA methods} &  \\ \hline

TFormer~\cite{sota_tformer} &90.10\\
GeSeNet~\cite{sota_gesenet}   &86.67\\
FusionM4Net~\cite{sota_fusionm4net} & 88.42\\

\hline

\end{tabular}
\end{center}
\end{table} 

\begin{table}
\small
%\tablestyle
\caption{Performance of different classifiers on the FLIR ATR dataset.} 
\label{tab:different_class_flir-atr}
\begin{center}       
\begin{tabular}{ll} %% this 
\hline
Classifier Name  & Accuracy (\%)\\ \hline
Single modality \footnotesize(MWIR) &84.29\\ 

Single modality {\footnotesize(LWIR)} &91.43\\ 

\textbf{FDCT }&\textbf{94.68}\\
\hline
{\footnotesize Comparison with other SOTA methods} &  \\ \hline

\textbf{TFormer}~\cite{sota_tformer} &\textbf{94.68}\\
GeSeNet~\cite{sota_gesenet}   &91.26\\
FusionM4Net~\cite{sota_fusionm4net} & 93.32\\

\hline

\end{tabular}
\end{center}
\end{table} 
\subsection{Experimental Results}
The proposed FDCT framework is experimented across four different multi-sensor ATR datasets, which exhibits better classification performance than the single modality classifier baseline. The comparative analysis, incorporating the confusion matrix and precision-recall curve, for both single modality and FDCT classifier's performances across four different datasets is illustrated in Figures~\ref{fig:Fig_combined_confusion} and ~\ref{fig:Fig_combined_PR}. Furthermore, we delve into the necessity of multi-modal alignment and decomposition losses. Subsequently, a comparative evaluation among the FDCT algorithm and other state-of-the-art multi-sensor fusion frameworks is conducted in Tables~\ref{tab:different_class_dsiac}, \ref{tab:different_class_VAIS}, \ref{tab:different_class_VEDAI}, and~\ref{tab:different_class_flir-atr}.

\subsubsection{Results on the DSIAC Dataset}
The DSIAC dataset comprises both mid-wave infrared and visible modalities ATR vehicles. We evaluate the classification performance of single modality classifiers by employing the ResNet-50~\cite{resnet} network. Notably, the MWIR and visible domains' single modality classifiers achieve an accuracy of 98.09\% and 98.67\%, respectively. Fusing these sensors' information within the FDCT algorithm, we achieve a classification accuracy of 99.29\%, as illustrated in Table~\ref{tab:different_class_dsiac}. The first row of Figures~\ref{fig:Fig_combined_confusion} and~\ref{fig:Fig_combined_PR} depicts the confusion matrices and precision-recall curves of the single modality classifiers and the FDCT framework on the DSIAC dataset. These visualizations underscore the superior performance of the fusion classifier (FDCT) across nearly every class compared to its single-modal counterparts.

\subsubsection{Results on the VAIS Maritime Dataset}
The VAIS dataset~\cite{VAIS_dataset} has pairs of infrared and visible maritime target chips. By assessing the performance of single-modality classifiers, we observe an accuracy of 88.00\% and 86.98\% within the infrared and visible modalities, respectively. The FDCT algorithm achieves superior performance, yielding a classification accuracy of 96.82\% compared to the single modality classifiers. The classifiers' performance on the VAIS dataset is illustrated in Table~\ref{tab:different_class_VAIS}. The confusion matrices and precision-recall curves, portraying the performance of the single modality classifiers alongside the FDCT classifier for the VAIS dataset, are depicted in the second row of Figures~\ref{fig:Fig_combined_confusion} and~\ref{fig:Fig_combined_PR}, respectively. This comprehensive visualization shows that the our framework outperforms single modality classifiers across all classes.

\subsubsection{Results on the VEDAI Aerial Dataset}
In the VEDAI dataset~\cite{VEDAI-dataset}, the single-modality classifiers secure accuracies of 86.11\% and 82.29\% within the infrared and visible modalities, respectively. Additionally, the proposed FDCT algorithm demonstrates superior performance, achieving an accuracy of 97.97\%, compared to single modality classifiers in Table~\ref{tab:different_class_VEDAI}. Notably, the enhancement provided by the FDCT classifier over single modality classifiers on the VEDAI dataset is substantial, particularly in the `Truck', `Tractor', `Van', `Others', and `Boat' classes. The third row of Figures~\ref{fig:Fig_combined_confusion} and~\ref{fig:Fig_combined_PR} demonstrates the confusion matrices and precision-recall curves of single modality and FDCT classifiers on the VEDAI dataset.

\subsubsection{Results on FLIR ATR dataset}
In the FLIR ATR dataset~\cite{asif_mehmood1}, the single modality classifiers secure an accuracy of 84.29\% and 91.43\% within the MWIR and LWIR modalities, respectively. Our proposed FDCT framework demonstrates superior performance, achieving an accuracy of 94.68\%, compared to single modality classifiers in Table~\ref{tab:different_class_flir-atr}. The classification performance of each class for both single modality and fusion (FDCT) classifiers on the FLIR ATR dataset is illustrated in the final row of Figures~\ref{fig:Fig_combined_confusion} and~\ref{fig:Fig_combined_PR}.

\subsubsection{Comparison with the SOTA methods} We compare the classification performance of our FDCT algorithm with the TFormer~\cite{sota_tformer}, FusionM4Net~\cite{sota_fusionm4net}, and GeSeNet~\cite{sota_gesenet}. While the TFormer and FusionM4Net represent fusion-based image classification methods, GeSeNet is a fusion-based image enhancement network. To attain the fused image classification performance in GeSeNet, we adopt its fusion module but substitute its decoder with a global average pooling and an MLP. Tables~\ref{tab:different_class_dsiac}, \ref{tab:different_class_VAIS}, \ref{tab:different_class_VEDAI}, and~\ref{tab:different_class_flir-atr} provide a comprehensive comparison, demonstrating the superior efficacy of our proposed FDCT method over TFormer, GeSeNet, and FusionM4Net in multi-sensor ATR target classification across the four datasets.

\subsubsection{Ablation Study}
\begin{table}[t]
    \caption{Ablation study of classification performance on the VEDAI dataset.}
    \centering
   % \resizebox{0.7\linewidth}{!}{
\begin{tabular}{c c c c c | c }
        \hline
        &\multicolumn{4}{c}{Losses in the training task}  &\\
        \hline
        &ITA & SCMA & CPA & Decomp. & Accuracy (\%) \\
        \hline
        
        &\xmark &\cmark &\cmark &\cmark & 97.35 \\
        %  & \cmark &\xmark \\
        % \cmark & \xmark & & 88.5 & 89.0 & & 88.8 & 89.6 & 90.5\\
        &\cmark & \xmark &\cmark &\cmark  & 96.90 \\
        &\cmark &\cmark & \xmark  &\cmark & 96.11 \\
        &\cmark & \cmark & \cmark &\xmark & 97.49\\ 
        &\cmark & \cmark & \cmark &\cmark & \textbf{97.97} 
        \\
        
        \hline
\end{tabular}\label{tab:ablation_study}
    
    \vspace{-0.3cm}
\end{table}
In the ablation study, we aim to scrutinize the significance of each loss function within the overall objective, discerning the effect of their removal from the total objective~(Eq.~\ref{eq:total_obj}). This is depicted in Table~\ref{tab:ablation_study}. This comparative evaluation of diverse objectives are evaluated on the VEDAI dataset, where the overall objective also includes the classifier's cross-entropy loss.\\

\noindent\textbf{Effect of Instance-wise Cross-Modal Alignment (ITA):}
From  Table~\ref{tab:ablation_study}, we can deduce that the ITA objective improves the classification performance by 0.62\%. Therefore, the ITA objective is an important component in the proposed FDCT algorithm.\\
\noindent\textbf{Effect of Sparse Cross-Modal Alignment (SCMA):}
The removal of the SCMA objective decreases the classification performance by 1.07\%. However, taking out the CPA objective causes a drop in classification accuracy by 1.86\%. Therefore, the CPA and SCMA are the first and second most important losses in the total objective.\\
\noindent\textbf{Effect of Cross-modal Prototype-Level Alignment (CPA):}
Our experiment illustrates that the CPA objective is more important than the ITA and SCMA objectives because the CPA objective regulates the alignment procedure among the heterogeneous UDT tokens. By removing the CPA objective, classification performance dropped from $97.97 \rightarrow 96.11$.\\ 
\noindent\textbf{Effect of Decomposition Loss:}
Incorporating the decomposition loss elevates the performance of our FDCT framework from $97.49 \rightarrow 97.97$. This loss improves the classification performance, which is relatively moderate, and it bolsters the extraction of meaningful and discriminative domain-specific and domain-invariant features.
\subsection{Future Directions}
We utilize the sparsemax to induce sparsity within our attention mechanisms in our proposed sparse cross-modal alignment scheme. However, Entmax~\cite{entmax}, fusedmax, and oscarmax~\cite{fusedmax} outperform sparsemax. In the future, researchers can investigate the performance of the fusion framework by utilizing other sparse attention in the alignment procedure. %Furthermore, in the unified discrete token (UDT) space, we utilize multi-head self-attention (MH{\text -}SA).
Moreover, others can investigate using a Swin transformer block~\cite{swin1} (instead of the vanilla transformer block) in the UDT space.
\section{Conclusion}\label{sec:conclusions}
This paper introduces our proposed FDCT framework to classify targets using visible and infrared sensor data by effectively extracting modality-dependent and modality-independent features. In response to the challenges posed by the heterogeneity between visible-infrared modalities, we introduce a shared unified discrete token space to reduce the granularity gap and an alignment module to subdue misalignment. Furthermore, we develop a sparse cross-modal alignment scheme using a cross-attention and sparsity constraint to capture fine-grained features between the modalities. Our proposed framework demonstrates superior performance in target classification compared to single-modality classifiers and several state-of-the-art multi-modal fusion algorithms. Finally, we expect the proposed alignment approach to be utilized in other multi-sensor fusion tasks, such as autonomous vehicle navigation, space and environmental observation, medical image fusion, etc.
\section*{Acknowledgement}
This research was funded by DEVCOM, Army Research Laboratory, and the contract number is W911NF2210117.
%%%%%%%%% REFERENCES

\bibliography{egbib}
%\bibliography{main}
\bibliographystyle{IEEEtran}

\begin{IEEEbiography}[{\includegraphics[width=1in,height=1.25in,clip,keepaspectratio]{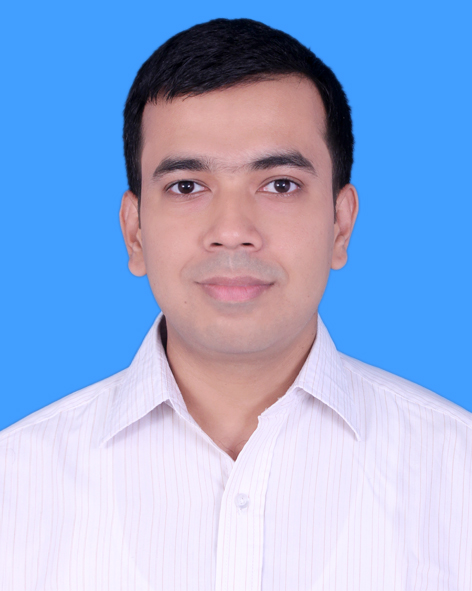}}]{Shoaib Meraj Sami} received the B.Sc. and the M.Sc. degree in Electrical and Electronic Engineering degree from the Bangladesh University of Engineering and Technology, Bangladesh, in 2016 and 2021, respectively. He also worked as an engineer at Nuclear Power Plant Company Bangladesh Limited and Chittagong Port Authority. He is also an author of \textsc{IEEE Transactions on Aerospace and Electronic Systems} and reviewer of  \textsc{IEEE Transactions on Automation Science and Engineering}. He is currently in his third year of a Ph.D at the Lane Department of Computer Science and Electrical Engineering, West Virginia University, Morgantown, WV, USA. His main focus is developing machine learning and deep learning algorithms and their applications with computer vision, biometrics, and automatic target recognition.
\end{IEEEbiography}

\begin{IEEEbiography}[{\includegraphics[width=1in,height=1.25in, clip,keepaspectratio]{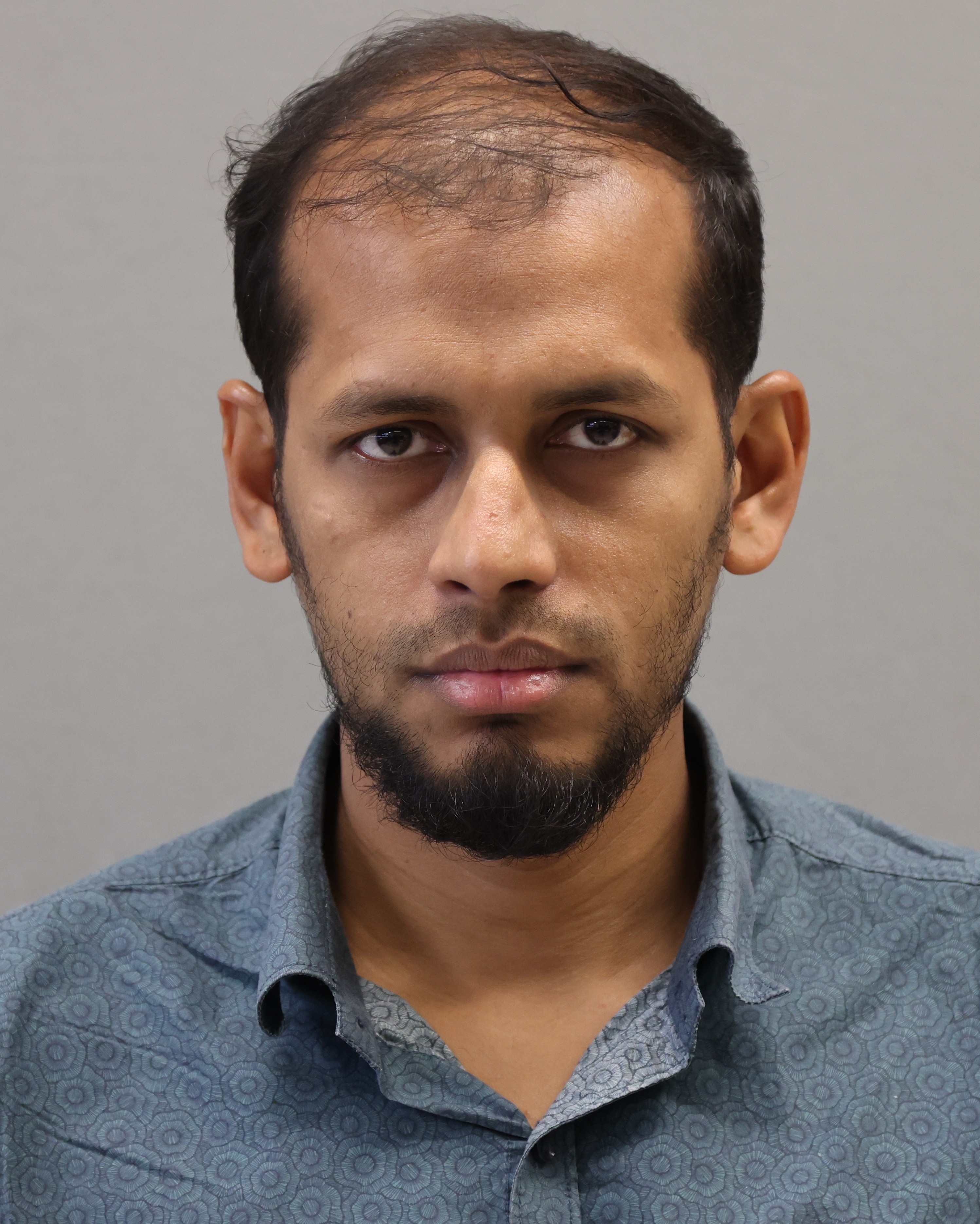}}] {Md Mahedi Hasan} is a Ph.D. student in the Lane Department of Computer Science and Electrical Engineering at West Virginia University in Morgantown, WV, USA. He received his B.Sc. in Electrical and Electronic Engineering from Khulna University of Engineering and Technology (KUET) and his M.Sc. in Information and Communication Technology from Bangladesh University of Engineering and Technology (BUET) in 2020. His research interests primarily focus on developing machine learning and deep learning algorithms, with an emphasis on their applications in computer vision and multimodal biometrics.
\end{IEEEbiography}

\begin{IEEEbiography}[{\includegraphics[width=1in,height=1.25in,clip,keepaspectratio]{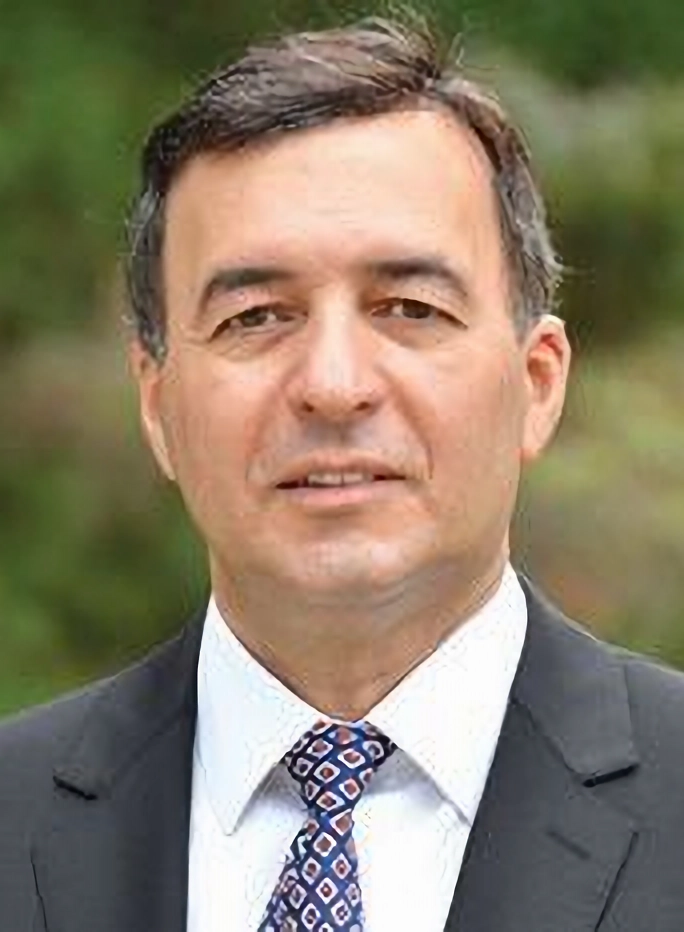}}]{Nasser M. Nasrabadi}(Fellow, IEEE) received the B.Sc. (Eng.) and Ph.D. degrees in
electrical engineering from the Imperial College
of Science and Technology, University of London,
London, U.K., in 1980 and 1984, respectively.
In 1984, he was at IBM, U.K., as a Senior Programmer. From 1985 to 1986, he was at the Philips
Research Laboratory, New York, NY, USA, as a
member of the Technical Staff. From 1986 to 1991,
he was an Assistant Professor at the Department
of Electrical Engineering, Worcester Polytechnic Institute, Worcester, MA,
USA. From 1991 to 1996, he was an Associate Professor at the Department
of Electrical and Computer Engineering, State University of New York at
Buffalo, Buffalo, NY, USA. From 1996 to 2015, he was a Senior Research
Scientist at the U.S. Army Research Laboratory. Since 2015, he has been
a Professor with the Lane Department of Computer Science and Electrical Engineering. His current research interests include image processing,
computer vision, biometrics, statistical machine learning theory, sparsity,
robotics, and deep neural networks. He is a fellow of the \textsc{International Society for Optical
Engineers (SPIE)}. He has served
as an Associate Editor for the \textsc{Ieee Transactions On Image Processing}, the
\textsc{IEEE Transactions On Circuits And Systems For Video Technology}, and the
\textsc{IEEE Transactions On Neural Networks And Learning Systems}.
\end{IEEEbiography}
\begin{IEEEbiography}[{\includegraphics[width=1in,height=1.25in,clip,keepaspectratio]{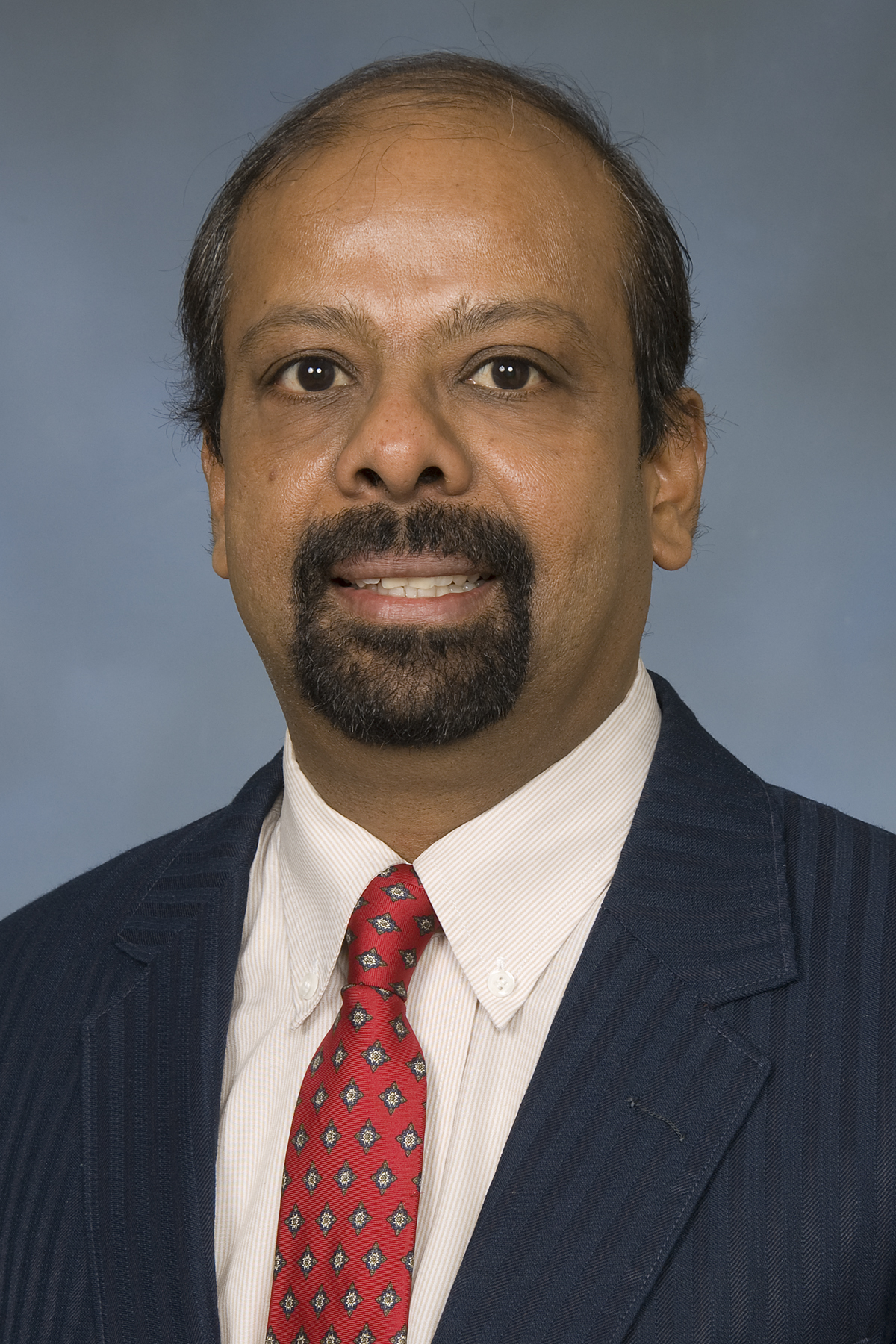}}]{Raghuveer M. Rao}(Fellow, IEEE)  received the M.E. degree in electrical communication engineering from
the Indian Institute of Science, Malleswaram, Bangalore, India, and the Ph.D. degree in electrical engineering from the University of Connecticut, Storrs,CT, USA, in 1981 and 1984, respectively.
He was a Member of Technical Staff at AMD
Inc., Santa Clara, CA, USA, from 1985 to 1987.
He joined the Rochester Institute of Technology,
Rochester, NY, USA, in December 1987, where,
at the time of leaving in 2008, he was a Professor of electrical engineering and imaging science. He is currently the Chief of the Intelligent Perception Branch, U.S. DEVCOM Army Research Laboratory (ARL), Adelphi, MD, USA, where he manages researchers and programs in computer vision and scene understanding. He has held visiting appointments at the Indian Institute of Science, the Air Force Research Laboratory, Dahlgren, VA, USA, the Naval Surface Warfare Center, Rome, NY, USA, and Princeton University, Princeton, NJ, USA. He has also served as an Associate Editor for the \textsc{Ieee
Transactions On Signal Processing, the Ieee Transactions On Circuits And Systems}, and \textsc{The Journal of Electronic Imaging}.
\end{IEEEbiography}
\end{document}